\documentclass[conference]{IEEEtran}
\IEEEoverridecommandlockouts
\usepackage{flushend}
\usepackage{cite}
\usepackage{amsmath,amssymb,amsfonts}
\usepackage{algorithmic}
\usepackage{graphicx}
\usepackage{textcomp}
\usepackage{xcolor}
\usepackage{cleveref} 
\usepackage{multirow}
\usepackage{xspace}
\usepackage{multicol}
\usepackage[utf8]{inputenc}
\usepackage{algorithm}
\usepackage{kotex}
\usepackage{xcolor}
\usepackage{booktabs}
\usepackage{multirow}
\usepackage{xspace}
\usepackage{microtype}
\usepackage{cite}
\usepackage{amsmath}
\usepackage{amssymb}
\usepackage{amsfonts}
\usepackage{balance}

\def\BibTeX{{\rm B\kern-.05em{\sc i\kern-.025em b}\kern-.08em
    T\kern-.1667em\lower.7ex\hbox{E}\kern-.125emX}}
\begin{document}
\title{Diffusion-Based Imitation Learning for Social Pose Generation}

\IEEEaftertitletext{\vspace{-2.5\baselineskip}}

%update the authors' details
\author{\IEEEauthorblockN{Antonio Martin-Ozimek}
\IEEEauthorblockA{
\textit{Honda Research Institute}\\
Wako, Japan \\
\textit{University of Alberta}\\
Edmonton, Canada \\
antonio2@ualberta.ca}%
\and
\IEEEauthorblockN{Isuru Jayarathne}
\IEEEauthorblockA{
\textit{Honda Research Institute}\\
Wako, Japan \\
isuru.jayarathne@jp.honda-ri.com}%
\and
\IEEEauthorblockN{Su Larb Mon}
\IEEEauthorblockA{
\textit{Honda Research Institute}\\
Wako, Japan \\
su.larbmon@jp.honda-ri.com}%
\and
\IEEEauthorblockN{Jouh Yeong Chew}
\IEEEauthorblockA{
\textit{Honda Research Institute}\\
Wako, Japan \\
jouhyeong.chew@jp.honda-ri.com}%
}%

 \makeatletter
\newcommand{\linebreakand}{%
  \end{@IEEEauthorhalign}
  \hfill\mbox{}\par
  \mbox{}\hfill\begin{@IEEEauthorhalign}
}
\makeatother

\IEEEaftertitletext{\vspace{-2.5\baselineskip}}
 
\maketitle
\begin{abstract}
Intelligent agents, such as robots and virtual agents, must understand the dynamics of complex social interactions to interact with humans. Effectively representing social dynamics is challenging because we require multi-modal, synchronized observations to understand a scene. We explore how using a single modality, the pose behavior, of multiple individuals in a social interaction can be used to generate nonverbal social cues for the facilitator of that interaction. The facilitator acts to make a social interaction proceed smoothly and is an essential role for intelligent agents to replicate in human-robot interactions. In this paper, we adapt an existing diffusion behavior cloning model to learn and replicate facilitator behaviors. Furthermore, we evaluate two representations of pose observations from a scene, one representation has pre-processing applied and one does not. The purpose of this paper is to introduce a new use for diffusion behavior cloning for pose generation in social interactions. The second is to understand the relationship between performance and computational load for generating social pose behavior using two different techniques for collecting scene observations. 
As such, we are essentially testing the effectiveness of two different types of conditioning for a diffusion model. We then evaluate the resulting generated behavior from each technique using quantitative measures such as mean per-joint position error (MPJPE), training time, and inference time. Additionally, we plot training and inference time against MPJPE to examine the trade-offs between efficiency and performance. Our results suggest that the further pre-processed data can successfully condition diffusion models to generate realistic social behavior, with reasonable trade-offs in accuracy and processing time. Future work will focus on extending this approach to generate multiple nonverbal social cues, on generalizing this method to multiple types of social activity, and on evaluating the results using human evaluators using a virtual agent or robot as a facilitator.
\end{abstract}


\section{Introduction}

There is growing interest in using robots and virtual agents to facilitate social interactions in real-world scenarios \cite{haru2018, chiang2024enhancing, liu2024peergpt, zhang2022storybuddy}. We propose that diffusion models are well-suited to tackling the intricate multi-modal challenge of modeling human behaviors in social interactions. Diffusion models have become widely recognized for their ability to adapt to various tasks, making them a powerful tool across many applications. These models use a probabilistic approach to handle complex and diverse data, allowing them to produce realistic and varied results even when the input is dynamic or unpredictable. This makes them particularly generalizable to a wide variety of tasks like behavior modeling, image creation, and generating social cues \cite{ho2020denoising, pearce2023imitating}.  Multi-modal data can also be easily integrated into diffusion models, which helps these models adapt to new situations \cite{saharia2022}. Recent improvements in optimization and sampling methods have also made them suitable for real-time use \cite{song2022}. As a result, diffusion models offer a promising approach for generating contextually aware outputs, while maintaining the ability to adapt to a wide variety of tasks and environments. 

One such use of these models is generating non-verbal social behaviors, such as gestures, gaze, and pose behavior; all of which are essential for social interaction facilitation \cite{ishi2021analysis,phutela2015importance}. For these systems to be truly effective, they must be able to adapt and generalize their behavior across various environments and social contexts. However, creating systems that can do this in real-time is a challenge, as it requires efficient models that can handle the variability and unpredictability of human behavior in different contexts \cite{chew2024joint}. In this paper, we aim to replicate the behavior of a conversation facilitator. A facilitator refers to a member of a social interaction who helps guide the interaction and help it proceed smoothly. The instructive work by Rosenberg et al. provides a clear example of how a robot can effectively facilitate class discussions\cite{rosenberg2020}. We build on their work by replicating the behavior of three different facilitators using behavior cloning. Since behavior cloning has always been trained on expert behavior. \cite{Bain1995AFF}, we decided to apply to our multi-facilitator use case.

While effective in controlled settings, behavior cloning (BC) often struggles to adapt to new or changing environments. This limitation arises because BC relies on supervised learning, mapping states to actions directly, without mechanisms for exploration or adaptation \cite{argall2009survey,levine2016}. When faced with scenarios outside the training data, small errors can quickly grow, leading to poor performance \cite{ross2011}. Combining BC with probabilistic models such as diffusion models has shown promise in making BC more robust and flexible for real-world applications \cite{chen2024diffaug}. As such, we want to apply the generalizability of a diffusion-based behavior cloning model to the novel scenario of generating social behavior for the facilitator of a group discussion .

In this work, we introduce a system designed to address the limitations of traditional behavior cloning by leveraging diffusion models. Building on the framework introduced in Imitating Human Behavior with Diffusion Models \cite{pearce2023imitating}, our system focuses on generating realistic human poses using changes in joint positions as actions (a) and equirectangular images as observations (o). This approach aims to capture the dynamics of social interactions while handling real-world variability. This model uses a transformer-based denoising network for the denoising process \cite{nichol2021improved}, which allows the system to learn patterns by gradually adding noise to the images, ensuring that the generated social behavior is both realistic and contextually relevant. Unlike traditional deterministic models, this system combines probabilistic modeling with behavior cloning to replicate the subtle nuances of human behavior in social settings.

Our key contribution is adapting diffusion models for social pose generation in a large group social interaction. Our secondary contribution is the evaluation of two different types of conditioning for our diffusion model. The first type of conditioning applies pre-processing to the images (o) and the second type uses the images themselves. In the following sections, we provide details about the dataset, describe the implementation of the diffusion-based behavior cloning system, and discuss the evaluation metrics used. We then present our results, explore the challenges of deploying the system in practical settings, and conclude with a summary of findings and future directions. This work underscores the importance of integrating diffusion-based methods into behavior cloning to create socially adept AI systems while assessing the most effective conditioning for these tasks.

\section{Method}

\begin{figure}[ht!]
    \centering
    \includegraphics[width=0.48\textwidth]{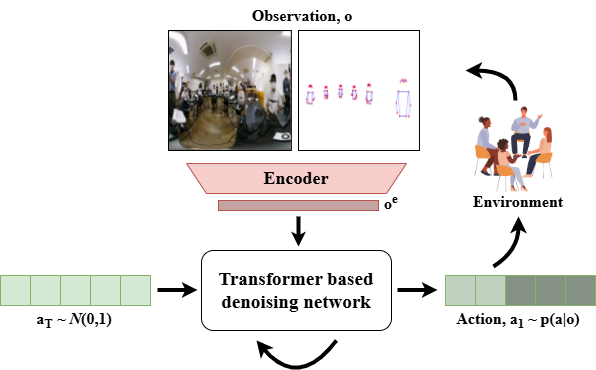} 
        \caption{Diffusion BC model flow chart}
        \label{fig:model}
\end{figure}

%% Added by Isuru related to DBC
\subsection{Model}
We adopted the approach introduced by Pearce et al. \cite{pearce2023imitating}, recognizing the inherent advantages of diffusion-based models over behavior cloning models\cite{chen2024diffaug}. This characteristic is particularly significant in social interaction scenarios, where the variability and unpredictability of responses can better emulate real-world dynamics. As detailed in their study, three distinct de-noising networks were evaluated: a basic multi-layer perceptron (MLP), an MLP Sieve, and a transformer model as seen in figure \ref{fig:model}. Among these, we selected the Transformer-based architecture for its superior accuracy during testing, making it the most reliable choice for our objectives. To generate the desired actions, we employed the Diffusion-X sampling technique, utilizing 50 de-noising time-steps $T$ and 8 refining steps $M$.
\subsubsection{Diffusion-X}
\begin{algorithm}
\caption{Diffusion-X Sampling Algorithm from Pearce et al. \cite{pearce2023imitating}} 
\label{algorithm}
\begin{algorithmic}[1]
\STATE $a_T \sim \mathcal{N}(0, I)$
\FOR{$i = T$ \TO $1 - M$}
    \STATE $\tau = \max(i, 1)$
    \IF{$\tau > 1$}
        \STATE $z \sim \mathcal{N}(0, I)$
    \ELSE
        \STATE $z = 0$
    \ENDIF
    \STATE $a_{\tau-1} = \frac{1}{\sqrt{\alpha_\tau}} \left(a_\tau - \frac{\sqrt{1-\alpha_\tau}}{\sqrt{1-\bar{\alpha}_\tau}} \theta(a_\tau, \tau, o) \right) + \sigma_\tau z$
\ENDFOR
\STATE \textbf{return} $a_{-M}$
\end{algorithmic}
\end{algorithm}
The algorithm represents a reverse diffusion process often used in generative models such as denoising diffusion probabilistic models (DDPM). Below, we explain its key components step by step:
\begin{itemize}
    \item \textbf{Initialization:} The algorithm begins by sampling $a_T$ from a standard normal distribution, $\mathcal{N}(0, I)$. This represents the initial noisy latent variable at the final timestep $T$.

    \item \textbf{Reverse Diffusion Process:} The process iterates backward from $T$ to $1 - M$, where $M$ is an offset value that determines the final step of the iteration.

    \item \textbf{Time Adjustment:} At each step $i$, the variable $\tau$ is computed as $\tau = \max(i, 1)$. This ensures that $\tau$ remains at least 1, preventing invalid indices.

    \item \textbf{Noise Sampling:} If $\tau > 1$, a noise variable $z$ is sampled from $\mathcal{N}(0, I)$. Otherwise, $z$ is set to 0. This condition ensures that no additional noise is introduced in the final step of the process.

    \item \textbf{Latent Variable Update:} The updated latent variable $a_{\tau-1}$ is computed using Algorithm \ref{algorithm}.
    \begin{itemize}
        \item $\alpha_\tau$ and $\bar{\alpha}_\tau$ are schedule coefficients controlling the amount of noise removed at each timestep.
        \item $\theta(a_\tau, \tau, o)$ is a model function that predicts the noise present in $a_\tau$ given the timestep $\tau$ and optional conditioning information $o$.
        \item $\sigma_\tau$ is the standard deviation of the noise added at timestep $\tau$, which accounts for stochasticity in the reverse process.
    \end{itemize}
\end{itemize}

\subsection{Multiparty Facilitation Data}
To train any behavior cloning model, we need to represent our data as a set of action (a) and observation (o) pairs\cite{schaal1996learning}. 

\begin{equation}
 D =\{ (o_1, a_1), (o_2, a_2)...  (o_n, a_n)\}
\end{equation}

To collect our data we utilized a subset of the FUMI-MPF \cite{chew2023teach}, which was derived from an experimental framework focused on group social interactions involving three distinct facilitator types. This multi-modal dataset was gathered during group discussions conducted under varying group configurations to explore diverse social dynamics. For this study, we selected three sessions for each facilitator type: musician, music teacher, and teacher. Each session consisted of 5 participants and 1 facilitator type from the three previously mentioned. As our model relies on images as observations from the environment, we used image data captured by a Theta Z1 360-degree camera.
To evaluate processing performance, we compared two distinct approaches: (1) using the raw 360-degree images, and (2) generating images by plotting extracted pose key points on a white background as seen in figure \ref{fig:img_type}. This comparison allowed us to assess the trade-offs between raw image data and pose-derived visual representations in terms of computational efficiency and model performance.

\begin{figure}[ht!]
    \centering
    \includegraphics[width=0.48\textwidth]{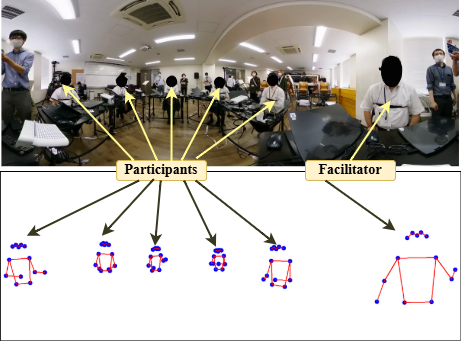} % Replace 'example-image' with your file name
        \caption{Two types of image used to test model performance}
        \label{fig:img_type}
\end{figure}

The frame rate of the Theta Z1 camera is about 30fps. For each frame, we extracted pose key points for all 6 participants. To prepare the training set, We calculated the differences between each consecutive frame of the $x$, $y$ coordinates of each of the facilitator's joints. We fed these $\Delta x$ and $\Delta y$ values into the model as actions ($a$). We resized the original 1920x960 pixel images to 128x128 pixel images in order to feed them to the model as observations ($o$). We used an 80-20 train-eval split for training our model with a total of about 90000 frames. For our research, we used an NVIDIA A100 80GB GPU.

\subsubsection{Evaluation Metrics}
To evaluate the accuracy of our model in generating human behavior, our proposed system incorporates the Mean Per Joint Position Error (MPJPE) as a key evaluation metric. MPJPE is widely used in human pose estimation to assess the precision of predicted poses. It measures the average Euclidean distance between the predicted joint positions and the ground truth joint positions. Since MPJPE quantifies the difference between the prediction and the actual pose, lower values of MPJPE indicate better performance, with an ideal value of 0 signifying perfect alignment.
This metric is particularly effective for evaluating generated poses, as it provides a quantitative assessment of the model’s ability to replicate human behavior. By employing MPJPE, we can directly compare the predicted poses from our diffusion-based model with the ground truth data, enabling us to analyze how closely the generated poses match the expected movements. Additionally, MPJPE is especially useful in this context because it allows us to assess not only individual keypoints but also overall pose accuracy, offering a holistic view of the model's performance in generating human-like social interactions. This approach aligns with established practices in pose estimation and generation tasks \cite{WANG2021103225}.

\section{Results}
\begin{figure}[h!]
    \centering
    \includegraphics[width=0.48\textwidth]{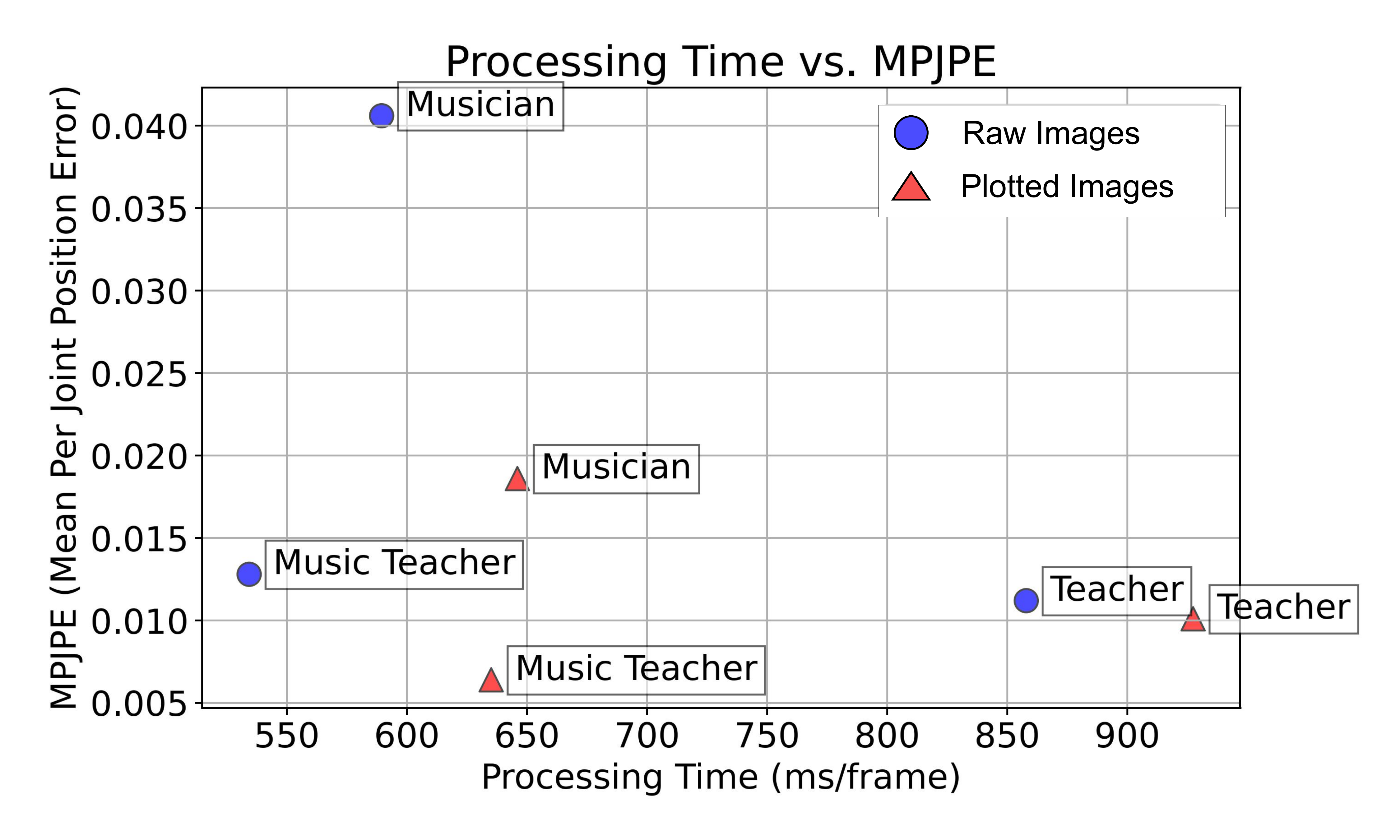} % Adjust the filename to your image file
    \caption{Processing Time vs. MPJPE for Raw Images and Plotted Images}
    \label{fig:time_vs_mpjpe}
\end{figure}
\begin{table}[t]
\normalsize
\caption{MPJPE Analysis of Predicted Poses (Lower is Better)}
\centering
\label{tab:mpjpe-analysis-result}
\begin{tabular}{@{}cccc@{}}
\toprule
\textbf{Facilitator} & \textbf{Raw Images} & \textbf{Plotted Images} & \textbf{\% Diff.} \\
\midrule
Teacher        & \textbf{0.0112}      & 0.0101      & 9.82\%
\\
Musician     & 0.0406     & 0.0186     & \textbf{54.19\%}
\\
Music Teacher   & 0.0128     & \textbf{0.0064}     & 50\%   
\\
\bottomrule
\end{tabular}
\end{table}
\begin{table}[ht]
\normalsize
\centering
\caption{Distribution Summary for RWrist Displacement of Facilitators}
\begin{tabular}{cccc}
\hline
\textbf{Subject} & \textbf{Mean} & \textbf{Standard Deviation} & \textbf{Range} \\
\hline
\text{Teacher} & 1.71 & 3.24 & 176.67 \\
\text{Musician} & 2.11 & 3.45 & 91.24 \\
\text{Music Teacher} & 3.18 & 5.52 & 114.21 \\
\hline
\end{tabular}

\label{tab:displacement_summary}
\end{table}

\begin{table}[h!]
\centering
\label{tab:processing_time}
\caption{Processing time and MPJPE for raw images and plotted images with percentage difference in processing time.}
\begin{tabular}{|c|c|c|c|c|}
\hline
\textbf{Subject} & \textbf{Dataset} & \textbf{Proc. Time (ms/frame)} & \textbf{MPJPE} \\
%\textbf{ } & \textbf{ } & \textbf{(ms/frame)} & \textbf{} \\
\hline
Teacher & Raw Images & 857.9208 & \textbf{0.0112}   \\
Musician & Raw Images & 589.5038 & 0.0406   \\
Music Teacher & Raw Images & \textbf{534.344} & 0.0128   \\
\hline
Teacher & Plot. Images & 927.397 \textbf{(+8.1)} & 0.0101  \\
Musician & Plot. Images & 646.032 \textbf{(+9.6)} & 0.0186  \\
Music Teacher & Plot. Images & \textbf{635.124} \textbf{(+18.9)} & \textbf{0.0064}  \\

\hline
\end{tabular}

\end{table}

Our evaluation metrics show how well the system replicates facilitator behavior across different sessions, with varying results depending on the context. The layout of Table \ref{tab:mpjpe-analysis-result} shows that the largest increase in MPJPE is seen from the Musician session at 54.19\% and is the smallest for the Teacher session at 9.82\%. The best performance on the raw images is the teacher facilitator and the best performance on the plotted images is the music teacher facilitator.
Looking at Table \ref{tab:displacement_summary} we see a representation of the distribution of the right wrist keypoint for the facilitator. We noticed that the right wrist and left wrist had similar distributions so we decided to present one of the two, the right wrist, in the discussion. Since our facilitator is seated during each discussion, we believe the wrists to be a good representation of non-verbal cues. We took the difference in positions of each consecutive frame and calculated the magnitude of the vector created to get an estimate of the movement generated by the non-verbal cues of the facilitator. We then figured out the characteristics of each of the distributions to get an idea of how large and how frequent these actions were. 
Finally, looking at Fig. \ref{fig:time_vs_mpjpe} we see the relationship between processing time and MPJPE. This value is a measure of the average time it takes to process a full frame from the pre-processing stage all the way through to the prediction stage. We see that the plotted images have better predictions along with higher processing times. 

\section{Discussion}

The primary topic is the increase in model performance when using our plotted images instead of the raw ones. As presented in Table \ref{tab:mpjpe-analysis-result}, we see an improvement across the board when using our plotted images. This improvement may stem from the pre-processing done to the image that extracts the necessary pose features and removes much of the noise that the normal image contains. Interestingly, there is not much of an improvement when looking at the Teacher session. Looking at the ground truth keypoints gives some insight into this phenomenon. The facilitator is seated during each session so we can look at the wrists as an indicator of the non-verbal cues generated during each session. We chose to look at the right wrist as its distributions were every similar to those of the left wrists. If we look at the distributions described in Table \ref{tab:displacement_summary}, we see that the mean and standard deviation of the teacher are similar to the values of the Musician while the standard deviation of the Music Teacher is quite unique. However, the range of the Teacher is quite a bit larger than the other two facilitators. We believe this is what may have cause the difference in prediction performance between the teacher and music teacher whose distributions seem quite similar in regards to mean and standard deviation. The larger exaggerated movements of the teacher were likely easier to pick up from the raw images than the subtle movements of the musician. Therefore, when we passed in the generated images, the model was able to better pick up on the subtle movements and predicted the musician's actions better. 

There seems to be no advantage with the plotted images in relation to the teacher as the movements are more easily seen in the raw images. The music teacher has the highest mean and standard deviation but does not have as large a range as the teacher. This causes the model to have a similar performance for both facilitators when using the raw image data because the movements can be said to be subtler. However, when we perform our feature extraction and generate new images, the model's performance on the music teacher increases noticeably as it is able to get a better understanding of these subtle movements.
While the model's performance does increase with our plotted images, the processing and computation time increases as well. Looking at Table \ref{tab:processing_time} we see that there is a stark increase in the processing time for each session. This mainly stems from the pre-processing where we generate the keypoints for all the participants in the raw image. The largest increase is in relation to the Music Teacher's session at about 18.9\%. We are not sure what caused this jump, however it is quite consistent with the processing time for the musician. The teacher has a significantly longer processing time than the other two sessions and we attribute this to background processes that were running on the GPU while during training. With the minimum processing time hovering at about 530ms/frame for the model trained on raw images and about 630ms/frame for the latter model, it is safe to say that neither implementation is ready for real-time implementation.

\section{Conclusion and Future Work}
 Seeing as the novelty of our approach relates to applying this model to a group social interaction by replicating a facilitator's actions, we plan to continue this branch of work by researching applications in other social settings. In discussing the applications of replicating the behavior of a group facilitator, we recognize the significant real-world potential of our research. Drawing inspiration from the works of S. Gillet et al. \cite{gillet2022learning} and R. Rosenberg-Kima et al. \cite{rosenberg2020}, we explore how intelligent systems, particularly robots, can enhance group learning interactions. S. Gillet’s research focuses on using a robot facilitator to balance engagement between a fluent speaker and a non-native learner in a language-learning context. Similarly, R. Rosenberg-Kima’s work highlights the utility of robots as teaching assistants. Her research places a robot in the role of a teaching assistant helping to manage lecture dynamics and improve learning outcomes. Building on this approach, we aim to expand the scope of our research to include larger, more diverse participant groups. Recognizing that there is much room for improvement, we plan to continue our work by optimizing the feature extraction and the diffusion model architecture to compensate for the increased processing time required for our implementations \cite{song2022}. We also recognize that the model needs to train on larger datasets to get a better understanding of its strengths and weaknesses. The results of this paper suggest that we need to test on more diverse social interactions.

\section*{ACKNOWLEDGMENT}
 This research is the result of a tripartite research collaboration between the University of Tokyo, Hiroo Gakuen Junior Senior High School, and Honda Research Institute Japan. The authors thank the collaborators for their assistance in implementing the experiments.

{
    \small
    \bibliographystyle{IEEEtran.bst}
    \bibliography{reference.bib}

\begin{thebibliography}{10}
\providecommand{\url}[1]{#1}
\csname url@samestyle\endcsname
\providecommand{\newblock}{\relax}
\providecommand{\bibinfo}[2]{#2}
\providecommand{\BIBentrySTDinterwordspacing}{\spaceskip=0pt\relax}
\providecommand{\BIBentryALTinterwordstretchfactor}{4}
\providecommand{\BIBentryALTinterwordspacing}{\spaceskip=\fontdimen2\font plus
\BIBentryALTinterwordstretchfactor\fontdimen3\font minus \fontdimen4\font\relax}
\providecommand{\BIBforeignlanguage}[2]{{%
\expandafter\ifx\csname l@#1\endcsname\relax
\typeout{** WARNING: IEEEtran.bst: No hyphenation pattern has been}%
\typeout{** loaded for the language `#1'. Using the pattern for}%
\typeout{** the default language instead.}%
\else
\language=\csname l@#1\endcsname
\fi
#2}}
\providecommand{\BIBdecl}{\relax}
\BIBdecl

\bibitem{haru2018}
\BIBentryALTinterwordspacing
R.~Gomez, D.~Szapiro, K.~Galindo, and K.~Nakamura, ``Haru: Hardware design of an experimental tabletop robot assistant,'' in \emph{Proceedings of the 2018 ACM/IEEE International Conference on Human-Robot Interaction}, ser. HRI '18.\hskip 1em plus 0.5em minus 0.4em\relax New York, NY, USA: Association for Computing Machinery, 2018, p. 233–240. [Online]. Available: \url{https://doi.org/10.1145/3171221.3171288}
\BIBentrySTDinterwordspacing

\bibitem{chiang2024enhancing}
C.-W. Chiang, Z.~Lu, Z.~Li, and M.~Yin, ``Enhancing ai-assisted group decision making through llm-powered devil's advocate,'' in \emph{Proceedings of the 29th International Conference on Intelligent User Interfaces}, 2024, pp. 103--119.

\bibitem{liu2024peergpt}
J.~Liu, Y.~Yao, P.~An, and Q.~Wang, ``Peergpt: Probing the roles of llm-based peer agents as team moderators and participants in children's collaborative learning,'' in \emph{Extended Abstracts of the CHI Conference on Human Factors in Computing Systems}, 2024, pp. 1--6.

\bibitem{zhang2022storybuddy}
Z.~Zhang, Y.~Xu, Y.~Wang, B.~Yao, D.~Ritchie, T.~Wu, M.~Yu, D.~Wang, and T.~J.-J. Li, ``Storybuddy: A human-ai collaborative chatbot for parent-child interactive storytelling with flexible parental involvement,'' in \emph{Proceedings of the 2022 CHI Conference on Human Factors in Computing Systems}, 2022, pp. 1--21.

\bibitem{ho2020denoising}
J.~Ho, A.~Jain, and P.~Abbeel, ``Denoising diffusion probabilistic models,'' \emph{Advances in neural information processing systems}, vol.~33, pp. 6840--6851, 2020.

\bibitem{pearce2023imitating}
T.~Pearce, T.~Rashid, A.~Kanervisto, D.~Bignell, M.~Sun, R.~Georgescu, S.~V. Macua, S.~Z. Tan, I.~Momennejad, K.~Hofmann \emph{et~al.}, ``Imitating human behaviour with diffusion models,'' \emph{arXiv preprint arXiv:2301.10677}, 2023.

\bibitem{saharia2022}
\BIBentryALTinterwordspacing
C.~Saharia, W.~Chan, S.~Saxena, L.~Li, J.~Whang, E.~Denton, S.~K.~S. Ghasemipour, B.~K. Ayan, S.~S. Mahdavi, R.~G. Lopes, T.~Salimans, J.~Ho, D.~J. Fleet, and M.~Norouzi, ``Photorealistic text-to-image diffusion models with deep language understanding,'' 2022. [Online]. Available: \url{https://arxiv.org/abs/2205.11487}
\BIBentrySTDinterwordspacing

\bibitem{song2022}
\BIBentryALTinterwordspacing
J.~Song, C.~Meng, and S.~Ermon, ``Denoising diffusion implicit models,'' 2022. [Online]. Available: \url{https://arxiv.org/abs/2010.02502}
\BIBentrySTDinterwordspacing

\bibitem{ishi2021analysis}
C.~T. Ishi and T.~Shintani, ``Analysis of eye gaze reasons and gaze aversions during three-party conversations.'' in \emph{Interspeech}, 2021, pp. 1972--1976.

\bibitem{phutela2015importance}
D.~Phutela, ``The importance of non-verbal communication,'' \emph{IUP Journal of Soft Skills}, vol.~9, no.~4, p.~43, 2015.

\bibitem{chew2024joint}
J.~Y. Chew and X.~Wang, ``Joint attention estimation during multi-party facilitation using multi-modal fusion,'' in \emph{Companion of the 2024 ACM/IEEE International Conference on Human-Robot Interaction}, 2024, pp. 322--326.

\bibitem{rosenberg2020}
R.~Rosenberg-Kima, Y.~Koren, and G.~Gordon, ``Robot-supported collaborative learning (rscl): Social robots as teaching assistants for higher education small group facilitation,'' \emph{Frontiers in Robotics and AI}, vol.~6, 01 2020.

\bibitem{Bain1995AFF}
\BIBentryALTinterwordspacing
M.~Bain and C.~Sammut, ``A framework for behavioural cloning,'' in \emph{Machine Intelligence 15}, 1995. [Online]. Available: \url{https://api.semanticscholar.org/CorpusID:10738655}
\BIBentrySTDinterwordspacing

\bibitem{argall2009survey}
B.~D. Argall, S.~Chernova, M.~Veloso, and B.~Browning, ``A survey of robot learning from demonstration,'' \emph{Robotics and autonomous systems}, vol.~57, no.~5, pp. 469--483, 2009.

\bibitem{levine2016}
\BIBentryALTinterwordspacing
S.~Levine, C.~Finn, T.~Darrell, and P.~Abbeel, ``End-to-end training of deep visuomotor policies,'' \emph{Journal of Machine Learning Research}, vol.~17, no.~39, pp. 1--40, 2016. [Online]. Available: \url{http://jmlr.org/papers/v17/15-522.html}
\BIBentrySTDinterwordspacing

\bibitem{ross2011}
\BIBentryALTinterwordspacing
S.~Ross, G.~J. Gordon, and J.~A. Bagnell, ``No-regret reductions for imitation learning and structured prediction,'' \emph{CoRR}, vol. abs/1011.0686, 2010. [Online]. Available: \url{http://arxiv.org/abs/1011.0686}
\BIBentrySTDinterwordspacing

\bibitem{chen2024diffaug}
\BIBentryALTinterwordspacing
S.-F. Chen, H.-C. Wang, M.-H. Hsu, C.-M. Lai, and S.-H. Sun, ``Diffusion model-augmented behavioral cloning,'' 2024. [Online]. Available: \url{https://arxiv.org/abs/2302.13335}
\BIBentrySTDinterwordspacing

\bibitem{nichol2021improved}
A.~Q. Nichol and P.~Dhariwal, ``Improved denoising diffusion probabilistic models,'' in \emph{International conference on machine learning}.\hskip 1em plus 0.5em minus 0.4em\relax PMLR, 2021, pp. 8162--8171.

\bibitem{schaal1996learning}
S.~Schaal, ``Learning from demonstration,'' \emph{Advances in neural information processing systems}, vol.~9, 1996.

\bibitem{chew2023teach}
J.~Y. Chew and K.~Nakamura, ``Who to teach a robot to facilitate multi-party social interactions?'' in \emph{Companion of the 2023 ACM/IEEE International Conference on Human-Robot Interaction}, 2023, pp. 127--131.

\bibitem{WANG2021103225}
\BIBentryALTinterwordspacing
J.~Wang, S.~Tan, X.~Zhen, S.~Xu, F.~Zheng, Z.~He, and L.~Shao, ``Deep 3d human pose estimation: A review,'' \emph{Computer Vision and Image Understanding}, vol. 210, p. 103225, 2021. [Online]. Available: \url{https://www.sciencedirect.com/science/article/pii/S1077314221000692}
\BIBentrySTDinterwordspacing

\bibitem{gillet2022learning}
S.~Gillet, M.~T. Parreira, M.~V{\'a}zquez, and I.~Leite, ``Learning gaze behaviors for balancing participation in group human-robot interactions,'' in \emph{2022 17th ACM/IEEE International Conference on Human-Robot Interaction (HRI)}.\hskip 1em plus 0.5em minus 0.4em\relax IEEE, 2022, pp. 265--274.

\end{thebibliography}
    \balance{}
    \bibliographystyle{unsrt}
    
%\columnbreak

}

\end{document}